\documentclass[10pt,twocolumn,letterpaper]{article}

\usepackage[pagenumbers]{cvpr} 

\usepackage{microtype}

\usepackage{algorithm}
\usepackage{algpseudocode}
\usepackage{array}
\usepackage{colortbl}
\usepackage{multirow}
\usepackage{stfloats}
\usepackage{tabularx}

\definecolor{fmaLightBlue}{RGB}{243,249,255}
\definecolor{fmaEmphasisBlue}{RGB}{205,227,255}

\definecolor{fmaLegendBlue}{RGB}{205,227,255}
\definecolor{fmaLegendGray}{RGB}{226,226,226}

\newcommand{\fmatablefont}{\fontsize{8}{9.6}\selectfont}
\newcolumntype{L}{>{\raggedright\arraybackslash}X}
\newcolumntype{Y}{>{\centering\arraybackslash}X}

\newsavebox{\fmatableshadebox}
\newlength{\fmatableshadewidth}
\newlength{\fmatableheaderheight}

\definecolor{cvprblue}{rgb}{0.21,0.49,0.74}
\usepackage[pagebackref,breaklinks,colorlinks,allcolors=cvprblue]{hyperref}
\DeclareRobustCommand{\FmaCrossText}[1]{\textcolor{cvprblue}{#1}}
\DeclareRobustCommand{\FmaCrossRef}[1]{\FmaCrossText{\ref*{#1}}}
\makeatletter
\DeclareRobustCommand{\FmaCrossEqref}[1]{\textup{\tagform@{\FmaCrossRef{#1}}}}
\makeatother

\def\confName{CVPR}
\def\confYear{2027}

\makeatletter
\patchcmd{\@maketitle}{\vspace*{24pt}}{\vspace*{6pt}}{}{}
\makeatother

\title{When Fog Becomes a Weapon: Atmospheric Attacks on Visible--Infrared Vision-Language Models}
\author{%
  Chengyin Hu\quad
  Dehe Zhao\quad
  Dingyi Lu\quad
  Qike Zhang\quad
  Jiaju Han\\
  Banglong Nong\quad
  Xu Kang\quad
  Jiujiang Guo\quad
  Kalibinuer Tiliwalidi\quad
  Jiahuan Long%
}
\hypersetup{
  pdftitle={When Fog Becomes a Weapon: Atmospheric Attacks on Visible--Infrared Vision-Language Models},
  pdfauthor={Chengyin Hu, Dehe Zhao, Dingyi Lu, Qike Zhang, Jiaju Han, Banglong Nong, Xu Kang, Jiujiang Guo, Kalibinuer Tiliwalidi, Jiahuan Long}
}

\begin{document}
\maketitle
\begin{abstract}
Existing adversarial attacks on vision-language models (VLMs) and cross-spectral perception primarily rely on conspicuous, unnatural artifacts, such as vivid adversarial patches or active light beams, that lack physical stealthiness, whereas prior weather attacks largely overlook how one atmosphere simultaneously affects heterogeneous sensors.
This raises a key scientific question: can a purely natural, shared fog environment simultaneously disrupt complementary visible and thermal infrared cues without artificial surface modifications? We introduce \textbf{Fog Manipulation Attack (FMA)}, a zero-gradient attack that optimizes a common 3D fog state to induce simultaneous semantic failures in visible-infrared VLMs.
FMA links shared, depth- and layer-aware atmospheric propagation along scene paths with band-specific imaging physics---visible airlight scattering and thermal path radiance---allowing one realistic fog state to erode both modalities' distinct cues.
Experiments show that FMA more than doubles simultaneous failure rates over fidelity-matched random fog, transfers to captioning and VQA, and removes cross-modal compensation, where restoring a clean companion view recovers 64\%--74\% of lost dual-input VQA answers.
\end{abstract}

\section{Introduction}\label{sec:intro}

Visible-infrared perception underpins robust multimodal intelligence because one spectral view can sustain recognition when the other degrades.
In vision-language models (VLMs)~\cite{li2023blip2,liu2024llava15,radford2021clip}, pairing visible (RGB) appearance with thermal infrared (IR) contrast provides strong resilience against poor illumination or clutter~\cite{vollmer2017infrared}.
However, existing adversarial attacks on VLMs and cross-spectral systems rely on conspicuous, unnatural artifacts---such as vivid patches~\cite{chen2026cfgpatch,hu2025touap,wei2023unified}, specialized garments~\cite{huang2026mecpatch,long2026thermal,zhu2026nonoverlapping}, or active light sources~\cite{hsiao2024naturallight,sayles2021light,zhu2021bulbs}---that lack physical stealthiness.

Fog offers a natural shared medium, yet ordinary fog rarely defeats dual-modal systems due to asymmetric modal degradation.
Atmospheric scattering and airlight~\cite{middleton1952vision,sakaridis2018foggy} can wash out texture until an object is obscured in RGB, while thermal emission still preserves clear boundaries in IR~\cite{vollmer2017infrared}; conversely, extinction and path radiance~\cite{vollmer2017infrared} can compress target-background contrast (e.g., a vehicle) in IR while object contours remain visible in RGB.
This raises a central question: \emph{can we optimize a cross-modal fog attack---governed by one shared atmosphere---that renders both RGB and IR observations unreliable simultaneously?} Prior weather attacks either target a single RGB modality~\cite{gao2021advhaze,fogfool2026} or optimize uncoupled perturbations, overlooking how one 3D atmosphere modulates both sensors through distinct radiative physics.

\begin{figure}[!t]
\centering
\includegraphics[width=\columnwidth,height=114.89285pt,keepaspectratio]{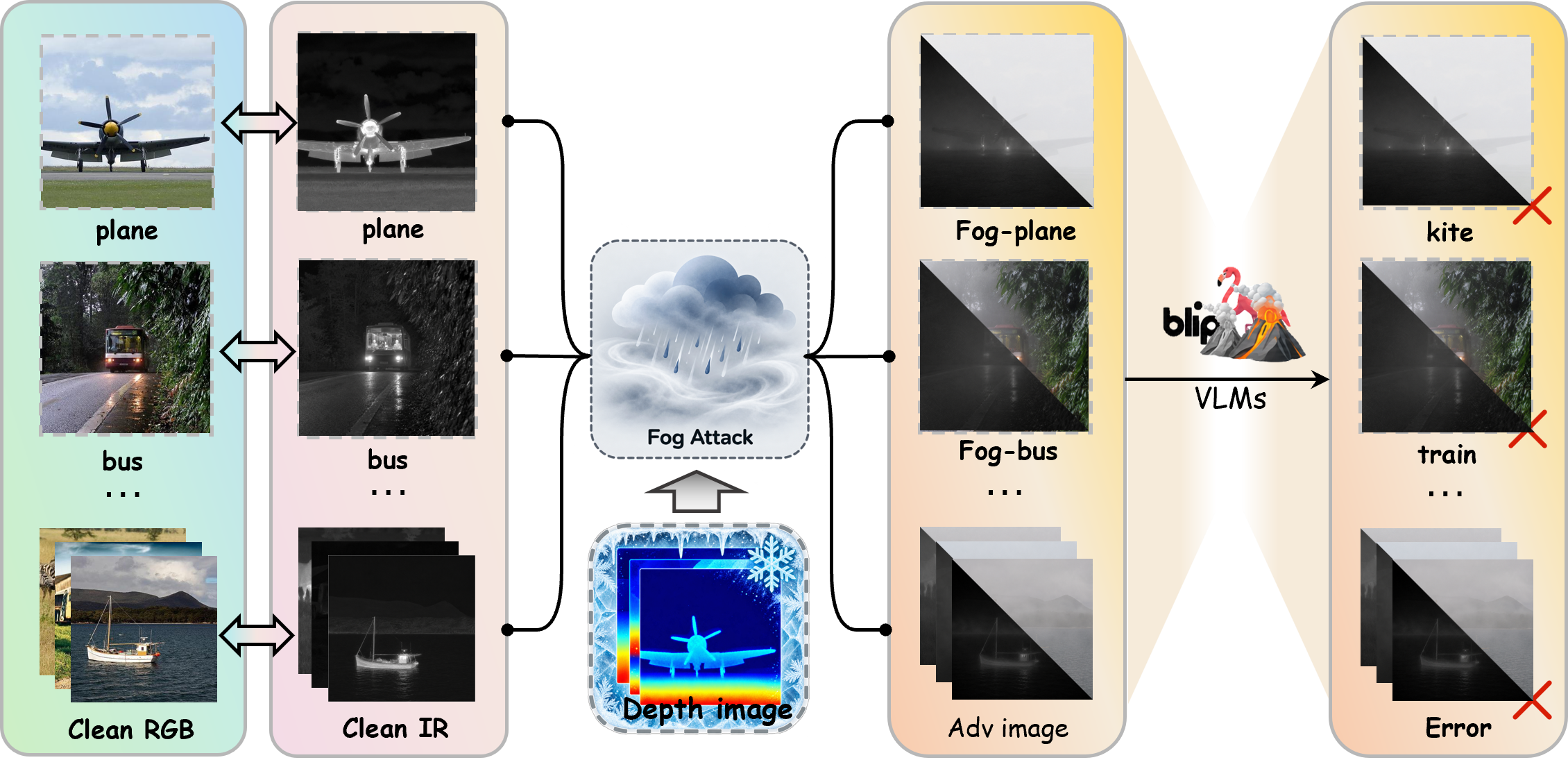}
\caption{\textbf{A shared fog attack on visible and thermal observations.}
FMA uses scene depth to render both views through one selected fog state.
The rightmost column illustrates semantic errors under paired fog, including predicting \textit{kite} for a plane and \textit{train} for a bus.}
\label{fig:intro_teaser}
\end{figure}

\begin{figure*}[!tbp]
\centering
\includegraphics[trim=5.455bp 0bp 6.327bp 0bp,clip,width=\linewidth]{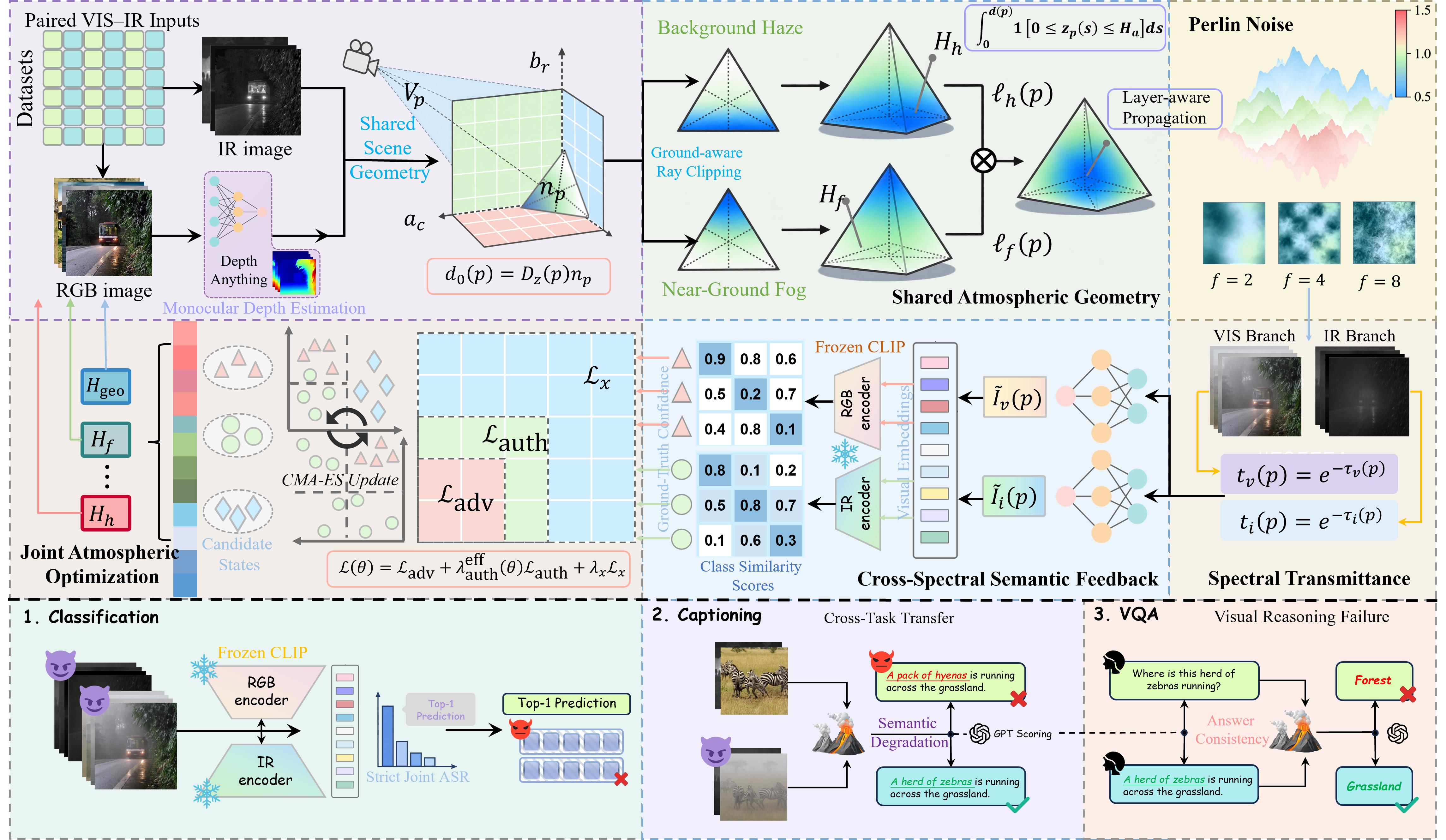}
\par
\caption{\textbf{Fog Manipulation Attack.}
FMA traces shared scene paths through near-ground fog and background haze, then renders visible and thermal observations with band-specific responses.
Scores from the two frozen classification branches guide the selection of a single fog state.
The selected pairs are evaluated on classification, captioning, and VQA.}
\label{fig:framework}
\end{figure*}

We introduce \textbf{Fog Manipulation Attack (FMA)} to bridge this gap (Figure~\ref{fig:intro_teaser}).
Starting from a registered RGB-IR pair, FMA reconstructs shared viewing rays via monocular depth~\cite{yang2024depthv2} and height-aware fog and haze layers~\cite{zhang2017fohis}, then renders both views through a compact 7D state coupling visible airlight scattering~\cite{middleton1952vision} with thermal extinction and path radiance~\cite{vollmer2017infrared}.
Guided solely by black-box scores from frozen RGB and IR branches, FMA searches for natural fog states that induce joint semantic failures while preserving structural fidelity~\cite{wang2004ssim}.

Our key contributions are:

\begin{itemize}

\item \textbf{Shared cross-spectral environmental attack.} We propose FMA, coupling unified 3D depth- and layer-aware atmospheric propagation with band-specific imaging physics so one realistic fog state disrupts both views simultaneously.

\item \textbf{Zero-gradient joint atmospheric search.} We design a compact 7D black-box search guided by dual-branch recognition feedback and fidelity regularization, transferring across unseen encoders and from classification to captioning and VQA.

\item \textbf{Cross-modal compensation breakdown.} Across controlled and real-sensor benchmarks, FMA more than doubles joint failure rates over fidelity-matched random fog and eliminates cross-modal compensation---restoring a clean companion view recovers 64\%--74\% of lost dual-input VQA answers.

\end{itemize}

\section{Related Work}\label{sec:related}

\subsubsection*{Attacking visual semantics.}

Vision-language models (VLMs) align visual encoders with text embeddings or language decoders for open-vocabulary classification, captioning, and visual question answering (VQA)~\cite{li2023blip2,liu2024llava15,radford2021clip}.
Because language outputs depend directly on visual tokens, perturbing the input image corrupts visual-semantic alignment---causing misclassifications, hallucinated captions, or flipped VQA answers.
Pixel- and feature-level attacks achieve this via generative perturbations (AnyAttack~\cite{zhang2025anyattack}) or value-feature manipulation (V-Attack~\cite{nie2026vattack}), while structured transformations alter global appearance via spatial warping~\cite{xiao2018spatial} or illumination~\cite{liu2025ita}.
Yet these single-stream RGB attacks leave the companion thermal view in dual-modal VLMs intact.

\subsubsection*{Attacking through the scene and weather.}

Physical attacks modify surfaces via patches and camouflage~\cite{brown2017patch,duan2020camouflage,eykholt2018robust,huang2020upc,zolfi2021translucent} or manipulate lighting~\cite{hsiao2024naturallight,sayles2021light,zhong2022shadows}, introducing conspicuous artifacts.
Weather benchmarks~\cite{kar20223dcc,sakaridis2018foggy} and sensor-fusion studies~\cite{bijelic2020seeing} show depth-dependent fog degrades perception, though passive weather rarely disables complementary sensors together.
Most closely related to FMA are adversarial weather attacks optimizing atmospheric parameters---from translated road fog~\cite{machiraju2020fog} and 3D particles~\cite{schmalfuss2023weather} to AdvHaze~\cite{gao2021advhaze} (depth-based scattering/airlight) and FogFool~\cite{fogfool2026} (2D Perlin fog masks).
However, these closest baselines remain confined to RGB images (or 2D masks lacking 3D geometry~\cite{fogfool2026}) and ignore thermal radiative transfer, leaving IR cues intact.

\subsubsection*{Infrared and cross-spectral attacks.}

Infrared attacks alter thermal signatures via heat sources, garments, or patches~\cite{wei2023irpatch,zhu2021bulbs,zhu2022clothing}, and recently target IR-adapted VLMs via grayscale patches (InfraPatch~\cite{hu2026infrapatch}) or thermal turbulence (AirflowAttack~\cite{su2026airflow}), but leave RGB cues intact.
Cross-spectral attacks must instead corrupt two sensors measuring distinct physical quantities---reflected visible light and emitted thermal radiation---through a single carrier.
Existing methods rely on localized surface or material coupling: Unified Patch~\cite{wei2023unified}, Adapted TOUAP~\cite{hu2025touap}, and CFGPatch~\cite{chen2026cfgpatch} share geometric contours across RGB and IR views (extending to VLM tasks~\cite{chen2026cfgpatch}),~\cite{zhu2026nonoverlapping} patterns non-overlapping RGB-T garments, and MEC-Patch~\cite{huang2026mecpatch} and~\cite{long2026thermal} couple reflectance with emissivity.
Unlike artificial surface modifications, FMA uses the shared 3D atmosphere itself as the coupling medium to disrupt both imaging processes simultaneously.

\microtypesetup{activate=false}

\section{Methodology}\label{sec:methodology}

FMA uses shared propagation paths to render a registered visible--thermal pair under one digital fog state, selected through paired semantic feedback (Figure~\ref{fig:framework}).

\subsection{Attacking a Pair through One Environment}\label{sec:threat_model}

The attacker starts with a visible image $I_v$, a thermal image $I_i$, and the target category $y$.
A frozen depth estimator provides scene geometry from the visible view.
The attacker can query category scores from frozen visible and IR classification surrogates; model gradients are unavailable.
Each search evaluation renders both images under one candidate state and queries the two branches.

A pair is eligible for Strict Joint attack success rate (ASR) when both branches recognize it correctly before the attack; an attack succeeds when both subsequently misclassify it.
For dual-input VQA, the corresponding readout tracks a correct unified answer becoming incorrect when the model receives both perturbed inputs.

\subsection{Building Shared Atmospheric Propagation}\label{sec:rendering}

\paragraph{Depth determines where fog accumulates.}
A spatially uniform overlay cannot express the different distances between a sensor, a nearby target, and the background.
FMA therefore uses a frozen Depth Anything V2 prediction \cite{yang2024depthv2} to construct viewing rays and approximate scene distances.
Downward rays end at the ground plane when that intersection occurs before the predicted surface.
Both spectral renderers share the above-ground paths to corresponding objects.

\paragraph{Layers determine which parts of a path are affected.}
Distance alone does not distinguish a low-lying fog layer from a deeper background haze.
Following FoHIS~\cite{zhang2017fohis}, FMA defines the layers by their upper heights and computes each ray's path length within them.
Changing a layer height can therefore change foreground and background attenuation differently, even when their image locations are close.
A fixed multi-scale Perlin field \cite{perlin1985noise} adds smooth spatial variation across rays.
The field is shared between the two views and held fixed during search; its multiplier is constant along each ray.

\paragraph{A compact state controls the environment.}
FMA searches a 7D state \(\boldsymbol{\theta}=(V_f,V_{\mathrm{aer}},H_f,H_h,V_{\mathrm{ir}},L_{\mathrm{th}},H_{\mathrm{geo}})\):
visible fog attenuation \(V_f\), shared background aerosol attenuation \(V_{\mathrm{aer}}\), near-ground fog height \(H_f\), background haze height \(H_h\), IR fog attenuation \(V_{\mathrm{ir}}\), thermal path radiance \(L_{\mathrm{th}}\), and vertical camera reference \(H_{\mathrm{geo}}\).
Layer heights, aerosol attenuation, and the geometry reference are common to the two branches.
The fog attenuation scales and thermal radiance let the spectral responses differ.
Parameter ranges, depth normalization, and the analytic layer intersections are specified in Supplementary Sec.~\FmaCrossRef{supp:renderer}.

\subsection{Rendering Different Spectral Responses}\label{sec:spectral}

The common propagation representation is converted into separate visible and thermal transmittance maps, $t_v$ and $t_i$.
Each map combines the shared path lengths and density with its band's fog attenuation; the background aerosol term is common.
The two images are then formed as
\begin{equation}
\begin{aligned}
\widetilde I_v(p)&=I_v(p)t_v(p)+A_v[1-t_v(p)],\\
\widetilde I_i(p)&=I_i(p)t_i(p)+L_{\mathrm{th}}[1-t_i(p)].
\end{aligned}
\label{eq:imaging}
\end{equation}
Here $p$ denotes a pixel, $A_v$ is visible airlight, and $L_{\mathrm{th}}$ is effective thermal path radiance in the normalized IR intensity scale.
The visible equation follows the atmospheric airlight model \cite{middleton1952vision}; the thermal equation models extinction and in-path radiance \cite{vollmer2017infrared}.
Both attenuate the original scene, but their added radiance changes contrast in different ways.
In particular, the thermal term allows the attack to alter radiative contrast that may remain informative after visible texture has faded.

\subsection{Searching for Joint Semantic Failure}\label{sec:objective}

The renderer turns the attack into a search over atmospheric states, denoted by $\boldsymbol\theta$.
FMA ranks candidates with three complementary signals:
\begin{equation}
\mathcal L(\boldsymbol\theta)=\mathcal L_{\mathrm{adv}}+
\lambda_{\mathrm{auth}}^{\mathrm{eff}}(\boldsymbol\theta)\mathcal L_{\mathrm{auth}}+\lambda_x\mathcal L_x.
\label{eq:total_loss}
\end{equation}
The semantic term $\mathcal L_{\mathrm{adv}}$ sums the two surrogates' probabilities for the correct category.
Minimizing it pushes both observations away from the original recognition result.
The fidelity term $\mathcal L_{\mathrm{auth}}$ penalizes the average loss of structural similarity to the clean images \cite{wang2004ssim}.
The transmittance term $\mathcal L_x$ penalizes disagreement between the two attenuation maps, discouraging candidates that concentrate the atmospheric effect in only one branch.

Fidelity receives greater weight once both branches have low correct-class confidence, favoring successful states with lower distortion.
Supplementary Sec.~\FmaCrossRef{supp:search} gives the component definitions and weighting rule.

\begin{figure*}[tbp]
\centering
\includegraphics[width=0.96009\linewidth,trim=7.8bp 0bp 0bp -3.39137bp,clip]{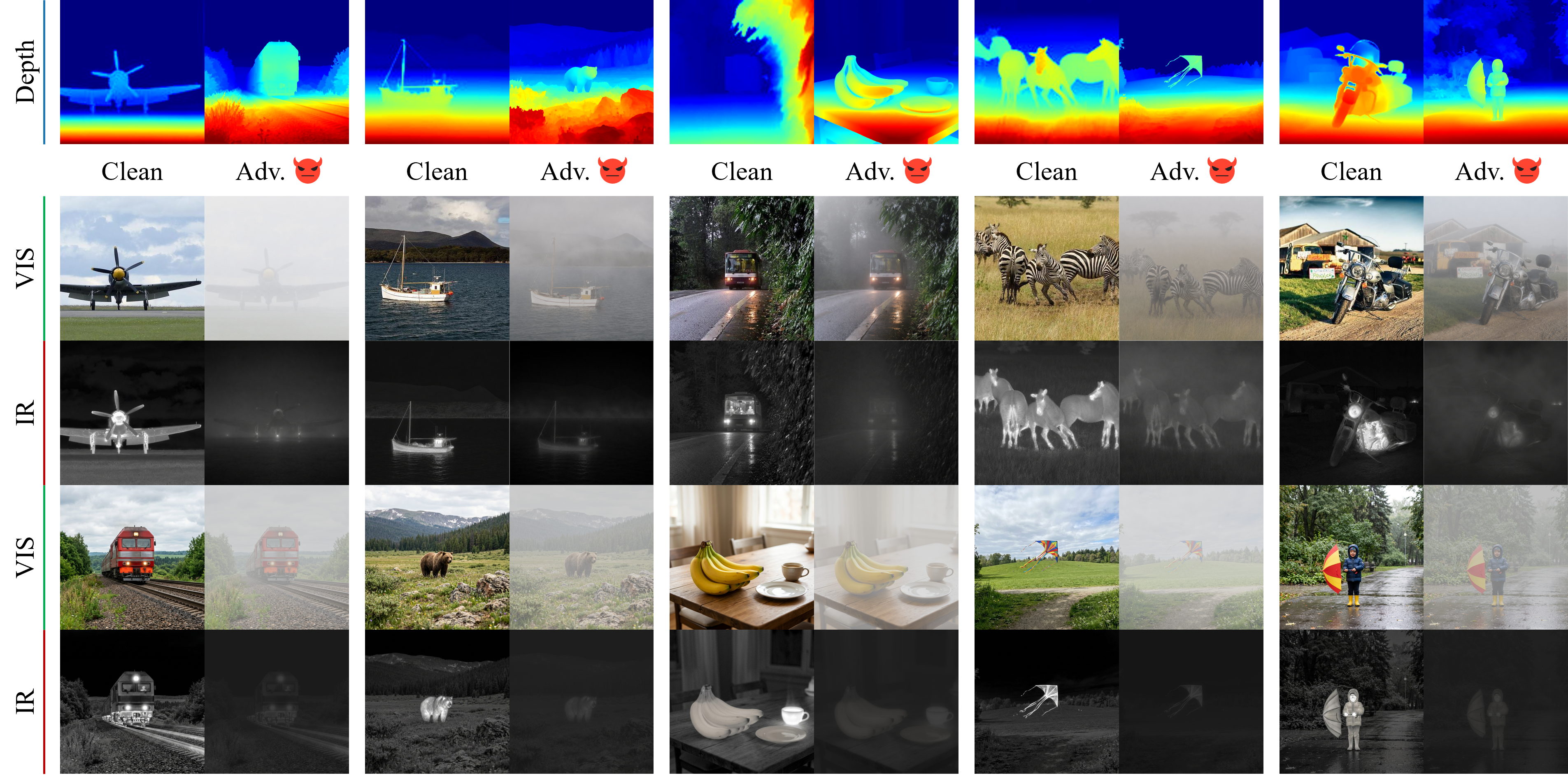}
\caption{RGB--IR attack examples. In the RGB and IR rows, clean observations are on the left and fog-perturbed observations on the right; the top row shows depth maps for the corresponding scenes.}
\label{fig:attack_examples}
\end{figure*}

\paragraph{Optimization and transfer.}\label{sec:optimization}
FMA uses CMA-ES \cite{hansen2001cmaes} to propose and rank atmospheric states from paired score queries.
Each candidate is decoded into the bounded atmospheric parameters, rendered in both bands, and evaluated with Eq.~\eqref{eq:total_loss}.
The search returns the lowest-objective candidate within the budget (Algorithm~\FmaCrossRef{alg:fma_planning}).
Depth, viewing rays, and density are prepared once per pair.
The renderer recomputes layer path lengths for each candidate.
Figure~\ref{fig:attack_examples} shows how a selected state changes visible detail and thermal contrast along the same scene paths.

After selection, the images are fixed for transfer tests, including captioning and VQA, so the downstream language outputs never guide the search.

\section{Experiments}\label{sec:experiments}

We test whether shared fog causes joint recognition failures and whether the disruption extends to captioning and VQA.

\subsection{Evaluation Setup}\label{sec:setup}

\paragraph{Datasets and tasks.}
The controlled benchmark pairs 300 MS COCO images across 30 categories~\cite{lin2014coco} with translated thermal views~\cite{lee2023thermal}, depth predictions, reference captions, and VQA questions~\cite{goyal2017vqav2}.
It supports comparison across classification, captioning, and question answering on the same scenes.
Real-sensor classification uses registered pairs from M3FD \cite{liu2022tardal}, FLIR-Aligned \cite{zhang2020cfr}, and FMB \cite{liu2023segmif}.
These datasets contribute 880 pairs spanning different sensors and capture conditions.
We also test naturally foggy DSERT-RoLL inputs \cite{cho2026dsert}, real-sensor VQA, and persistence across adjacent video frames.
Supplementary Secs.~\FmaCrossRef{supp:data} and~\FmaCrossRef{supp:scoring} detail sampling, target windows, annotations, and splits.

\paragraph{Models and attack selection.}
The classification surrogates are OpenAI CLIP ViT-L/14 \cite{radford2021clip}, OpenCLIP ViT-B/16 \cite{cherti2023openclip}, MetaCLIP ViT-L/14 \cite{xu2024metaclip}, and EVA-CLIP ViT-G/14 \cite{sun2023evaclip}, with frozen IR-adapted branches.
Single-stream language evaluation covers six models from the LLaVA, OpenFlamingo, BLIP-2, and InstructBLIP families \cite{liu2024llava15,awadalla2023openflamingo,li2023blip2,dai2023instructblip}.
Their source encoders are identified in Supplementary Tables~\FmaCrossRef{tab:s12} and~\FmaCrossRef{tab:s13}; the projectors and language decoders do not participate in attack selection.
Both views are passed to LLaVA-NeXT-Interleave~\cite{li2025llavainterleave} and Qwen2-VL~\cite{wang2024qwen2vl}, with attacks selected on MetaCLIP.
The default search uses 840 paired evaluations.
Models, adaptation, and exceptions are detailed in Supplementary Sec.~\FmaCrossRef{supp:models}.

\begin{table*}[!tbp]
\centering
\captionsetup{skip=6pt}
\caption{Classification ASR (\%) on the controlled benchmark.
Joint denotes Strict Joint ASR, and Macro averages the four encoders equally.
{\setlength{\fboxsep}{1pt}\protect\colorbox{fmaLegendBlue}{\strut Blue}} to
{\setlength{\fboxsep}{1pt}\protect\colorbox{fmaLegendGray}{\strut gray}}
indicates higher to lower ASR within each column, with the highest in bold.
Supplementary Table~\FmaCrossRef{tab:s1} gives the exact counts.}
\label{tab:m1}

\begingroup\fmatablefont
\colorlet{fmaTableOneGrade1}{fmaLegendGray}
\definecolor{fmaTableOneGrade2}{RGB}{238,238,238}
\definecolor{fmaTableOneGrade3}{RGB}{248,248,248}
\colorlet{fmaTableOneGrade4}{fmaLightBlue}
\definecolor{fmaTableOneGrade5}{RGB}{224,238,255}
\colorlet{fmaTableOneGrade6}{fmaLegendBlue}
\setlength{\tabcolsep}{2.2pt}
\renewcommand{\arraystretch}{0.92}
\begin{tabularx}{\linewidth}{@{}l*{13}{Y}}
\toprule
\multirow{2}{*}[-2pt]{Method}
 & \multicolumn{3}{c}{OpenAI CLIP ViT-L/14}
 & \multicolumn{3}{c}{OpenCLIP ViT-B/16}
 & \multicolumn{3}{c}{MetaCLIP ViT-L/14}
 & \multicolumn{3}{c}{EVA-CLIP ViT-G/14}
 & Macro \\
\cmidrule(lr){2-4}\cmidrule(lr){5-7}\cmidrule(lr){8-10}\cmidrule(lr){11-13}\cmidrule(lr){14-14}
 & VIS & IR & Joint & VIS & IR & Joint & VIS & IR & Joint & VIS & IR & Joint & Joint \\
\midrule
Random Fog (4D Match) & \cellcolor{fmaTableOneGrade1}17.12 & \cellcolor{fmaTableOneGrade2}46.57 & \cellcolor{fmaTableOneGrade1}10.50
 & \cellcolor{fmaTableOneGrade1}21.28 & \cellcolor{fmaTableOneGrade1}43.98 & \cellcolor{fmaTableOneGrade1}12.77
 & \cellcolor{fmaTableOneGrade2}17.81 & \cellcolor{fmaTableOneGrade3}41.15 & \cellcolor{fmaTableOneGrade1}8.51
 & \cellcolor{fmaTableOneGrade1}12.84 & \cellcolor{fmaTableOneGrade3}38.61 & \cellcolor{fmaTableOneGrade1}8.96
 & \cellcolor{fmaTableOneGrade1}10.18 \\
Unified Patch \cite{wei2023unified} & \cellcolor{fmaTableOneGrade1}20.21 & \cellcolor{fmaTableOneGrade1}34.80 & \cellcolor{fmaTableOneGrade2}13.00
 & \cellcolor{fmaTableOneGrade1}22.97 & \cellcolor{fmaTableOneGrade1}38.74 & \cellcolor{fmaTableOneGrade2}16.49
 & \cellcolor{fmaTableOneGrade1}11.30 & \cellcolor{fmaTableOneGrade1}15.10 & \cellcolor{fmaTableOneGrade1}8.51
 & \cellcolor{fmaTableOneGrade1}15.20 & \cellcolor{fmaTableOneGrade1}20.30 & \cellcolor{fmaTableOneGrade2}11.94
 & \cellcolor{fmaTableOneGrade2}12.49 \\
Adapted TOUAP \cite{hu2025touap} & \cellcolor{fmaTableOneGrade5}32.88 & \cellcolor{fmaTableOneGrade2}45.10 & \cellcolor{fmaTableOneGrade3}15.00
 & \cellcolor{fmaTableOneGrade2}26.01 & \cellcolor{fmaTableOneGrade2}49.74 & \cellcolor{fmaTableOneGrade3}20.21
 & \cellcolor{fmaTableOneGrade6}\textbf{38.01} & \cellcolor{fmaTableOneGrade3}40.10 & \cellcolor{fmaTableOneGrade3}13.30
 & \cellcolor{fmaTableOneGrade6}\textbf{35.47} & \cellcolor{fmaTableOneGrade5}51.49 & \cellcolor{fmaTableOneGrade5}17.41
 & \cellcolor{fmaTableOneGrade3}16.48 \\
CFGPatch \cite{chen2026cfgpatch} & \cellcolor{fmaTableOneGrade6}33.90 & \cellcolor{fmaTableOneGrade2}48.04 & \cellcolor{fmaTableOneGrade3}16.50
 & \cellcolor{fmaTableOneGrade2}28.04 & \cellcolor{fmaTableOneGrade3}52.88 & \cellcolor{fmaTableOneGrade4}21.81
 & \cellcolor{fmaTableOneGrade6}36.99 & \cellcolor{fmaTableOneGrade4}42.71 & \cellcolor{fmaTableOneGrade4}14.89
 & \cellcolor{fmaTableOneGrade6}34.46 & \cellcolor{fmaTableOneGrade5}52.97 & \cellcolor{fmaTableOneGrade5}18.41
 & \cellcolor{fmaTableOneGrade4}17.90 \\
AdvHaze (Depth-1L) \cite{gao2021advhaze} & \cellcolor{fmaTableOneGrade5}30.14 & \cellcolor{fmaTableOneGrade4}64.22 & \cellcolor{fmaTableOneGrade4}18.00
 & \cellcolor{fmaTableOneGrade4}35.14 & \cellcolor{fmaTableOneGrade5}63.87 & \cellcolor{fmaTableOneGrade4}22.87
 & \cellcolor{fmaTableOneGrade5}29.45 & \cellcolor{fmaTableOneGrade5}59.38 & \cellcolor{fmaTableOneGrade4}15.96
 & \cellcolor{fmaTableOneGrade3}22.97 & \cellcolor{fmaTableOneGrade5}55.94 & \cellcolor{fmaTableOneGrade4}15.92
 & \cellcolor{fmaTableOneGrade4}18.19 \\
FogFool (Perlin-2D) \cite{fogfool2026} & \cellcolor{fmaTableOneGrade4}28.08 & \cellcolor{fmaTableOneGrade4}60.78 & \cellcolor{fmaTableOneGrade3}16.50
 & \cellcolor{fmaTableOneGrade4}33.11 & \cellcolor{fmaTableOneGrade4}61.26 & \cellcolor{fmaTableOneGrade4}21.28
 & \cellcolor{fmaTableOneGrade4}27.40 & \cellcolor{fmaTableOneGrade5}56.25 & \cellcolor{fmaTableOneGrade4}14.89
 & \cellcolor{fmaTableOneGrade3}21.62 & \cellcolor{fmaTableOneGrade5}52.97 & \cellcolor{fmaTableOneGrade4}14.93
 & \cellcolor{fmaTableOneGrade4}16.90 \\
FMA (Ours) & \cellcolor{fmaTableOneGrade6}\textbf{36.30} & \cellcolor{fmaTableOneGrade6}\textbf{79.41} & \cellcolor{fmaTableOneGrade6}\textbf{24.00}
 & \cellcolor{fmaTableOneGrade6}\textbf{44.93} & \cellcolor{fmaTableOneGrade6}\textbf{75.39} & \cellcolor{fmaTableOneGrade6}\textbf{29.79}
 & \cellcolor{fmaTableOneGrade6}37.33 & \cellcolor{fmaTableOneGrade6}\textbf{70.31} & \cellcolor{fmaTableOneGrade6}\textbf{20.21}
 & \cellcolor{fmaTableOneGrade4}27.03 & \cellcolor{fmaTableOneGrade6}\textbf{65.84} & \cellcolor{fmaTableOneGrade6}\textbf{20.40}
 & \cellcolor{fmaTableOneGrade6}\textbf{23.60} \\
\bottomrule
\end{tabularx}
\endgroup
\end{table*}

\begin{figure*}[tbp]
\centering
\includegraphics[width=0.95277\linewidth,trim=10.5bp 0bp 0bp 8.7bp,clip]{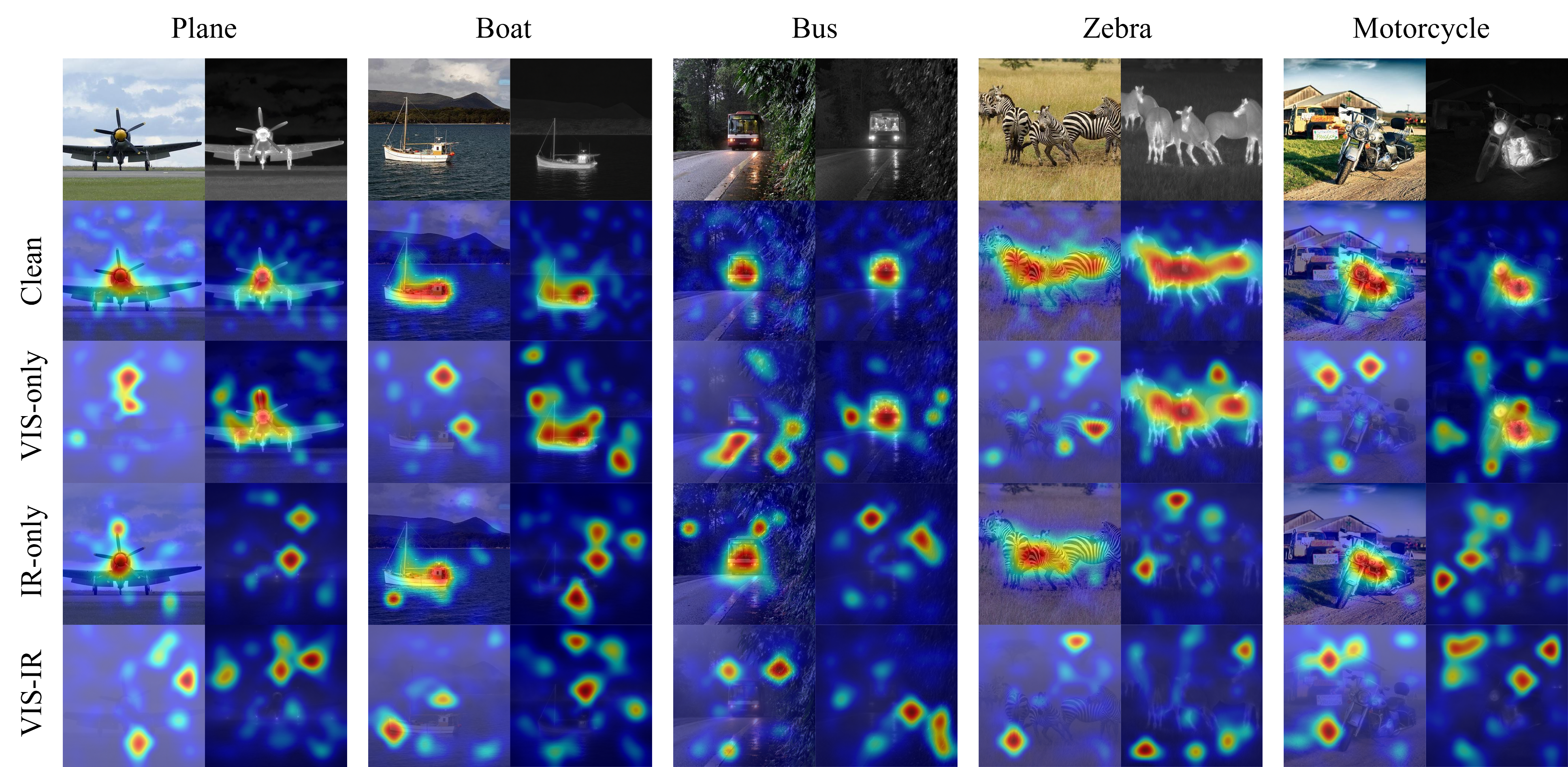}
\caption{Spatial response redistribution under single- and dual-branch perturbations.
Each column pairs VIS (left) and IR (right) views across Clean, VIS-only, IR-only, and Paired attack conditions, showing how paired fog disrupts target localization across both modalities.}
\label{fig:attention_redistribution}
\end{figure*}

\paragraph{Comparisons and readouts.}
We compare FMA with fidelity-matched Random Fog, optimized weather baselines, and cross-spectral patches.
Weather baselines include uniform fog, adapted AdvHaze, and adapted FogFool. Patch baselines include Unified Patch, Adapted TOUAP, and CFGPatch (Supplementary Sec.~\FmaCrossRef{supp:baselines}).
Random Fog matches the structural and perceptual distortion of the selected FMA pair to control for fog severity.

Classification reports branch ASR and Strict Joint ASR.
Captioning and VQA use a blinded scoring protocol with human-audit results in Supplementary Sec.~\FmaCrossRef{supp:scoring}.
Real-data uncertainty is estimated by resampling physical scene clusters while retaining paired observations across methods and encoders.

\subsection{Inducing Failure in Both Spectral Branches}\label{sec:controlled}

Table~\ref{tab:m1} shows that FMA has the highest IR and Strict Joint ASR on all four controlled-benchmark surrogates.
Mean Strict Joint ASR rises from 10.18\% under fidelity-matched Random Fog to 23.60\% under FMA, which also exceeds the optimized haze and patch baselines on this metric.
FMA and Random Fog use the same renderer at comparable structural and perceptual distortion.

Adapted TOUAP has the highest VIS ASR on some encoders, while FMA produces the most simultaneous failures.
Supplementary Table~\FmaCrossRef{tab:s1} gives the branch and joint counts behind these rates.

As visualized in Figure~\ref{fig:attention_redistribution}, perturbing a single branch leaves target-focused responses intact in the clean companion view, whereas FMA's paired perturbation disperses spatial focus away from the target in both modalities simultaneously.

\subsection{Recognition Errors and Compensation Loss}\label{sec:vlm}

\begin{figure}[!tbp]
\centering
\includegraphics[pagebox=cropbox,trim=0bp -2.90002bp 0bp 0bp,clip,width=\columnwidth]{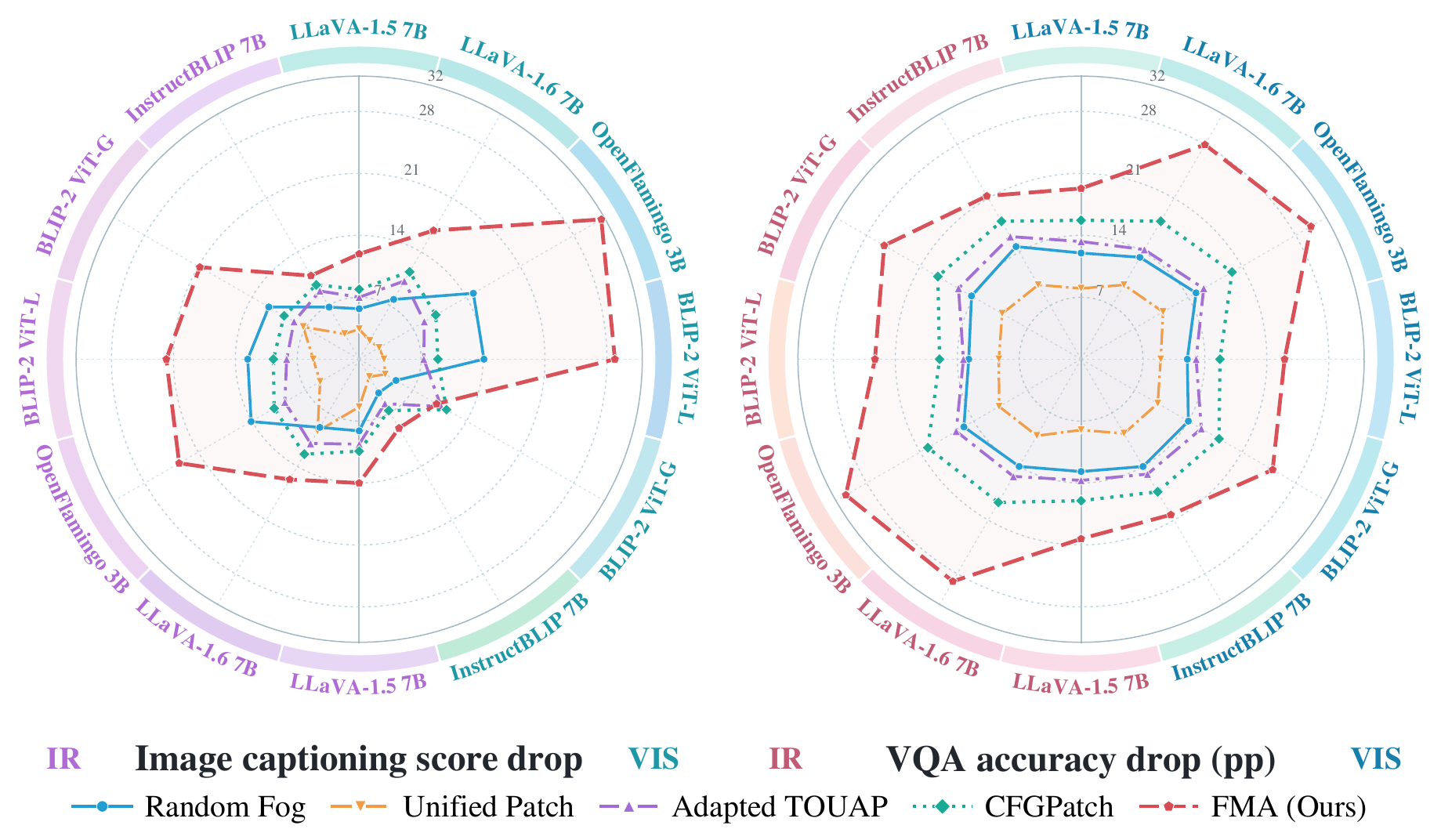}
\caption{Drops from clean inputs for single-stream image captioning (left, score points)
and VQA (right, percentage points) across six models per modality.
Supplementary Tables~\FmaCrossRef{tab:s12} (VQA) and~\FmaCrossRef{tab:s13} (captioning) report the full scores, drops, and source encoders.}
\label{fig:language_radar}
\end{figure}

\paragraph{Language outputs inherit the visual disruption.}
The classification-selected images degrade captioning across both spectral branches
(Figure~\ref{fig:language_radar}, left). Full scores and drops are in Supplementary Table~\FmaCrossRef{tab:s13}.
The lower caption score in each pair tracks the less informative view and falls on average from 63.40 on clean inputs to 45.12 under FMA.

Figure~\ref{fig:language_radar} (right) shows VQA accuracy drops in percentage points.
Supplementary Table~\FmaCrossRef{tab:s12} gives the full accuracies and drops.
Across the six single-stream VLMs, FMA causes both answers to become incorrect on 41.8\% of the pairs for which both were initially correct.
Random Fog gives 21.0\%.
Supplementary Table~\FmaCrossText{\FmaCrossRef{tab:s8}(b)} gives the answer-state transitions.

\begin{figure*}[bp]
\centering
\includegraphics[pagebox=cropbox,width=\linewidth]{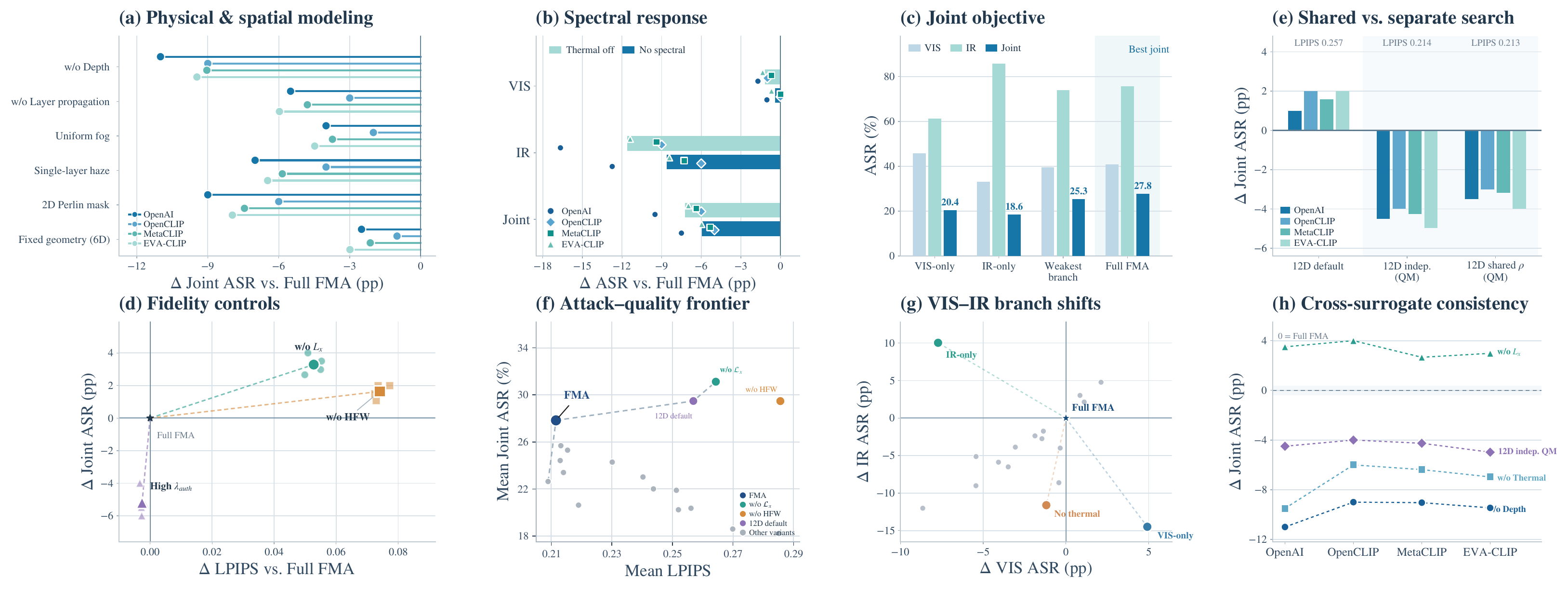}
\caption{\textbf{Design ablations on controlled pairs.}
Panels (a)--(e) vary spatial structure, spectral response, the objective, fidelity controls, and parameter sharing, respectively.
Panels (f)--(h) show the attack--fidelity frontier, VIS--IR branch shifts, and agreement across encoders.
Each search uses 1,260 paired evaluations. Deltas are relative to each encoder's Full FMA, and means weight the four encoders equally.
In (b), bars show means and markers show individual encoders.
QM denotes quality matching and HFW denotes hierarchical fidelity weighting.
Supplementary Sec.~\FmaCrossRef{supp:controlled_ablations} and Table~\FmaCrossRef{tab:s11} give the protocol and per-encoder results.}
\label{fig:controlled_ablations}
\end{figure*}

\begin{figure*}[tbp]
\centering
\includegraphics[width=0.90\linewidth]{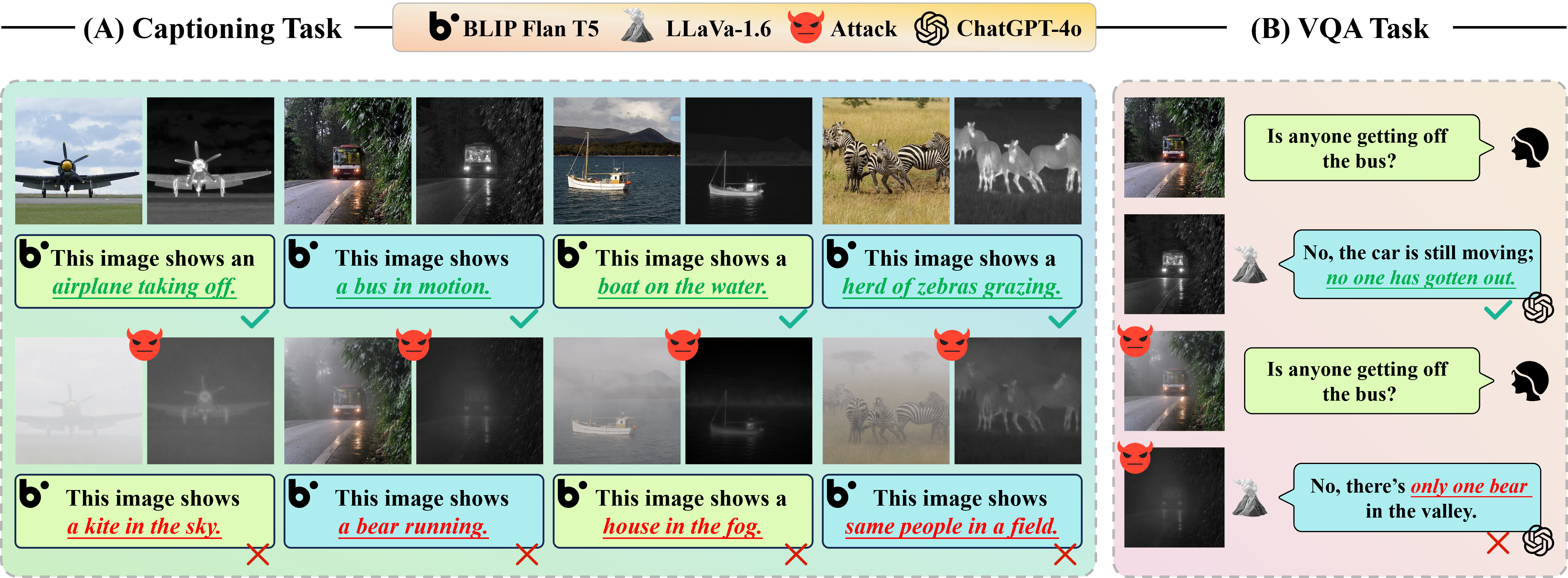}
\captionsetup{skip=3pt}
\caption{\textbf{Captioning (A) and VQA (B) under FMA.}
Each panel shows clean inputs and their outputs above the corresponding fogged examples.
Ticks and crosses mark the evaluator's judgments (Supplementary Sec.~\FmaCrossRef{supp:scoring}).}
\label{fig:qualitative_outputs}
\end{figure*}

\begin{figure}[!tbp]
\centering
\includegraphics[pagebox=cropbox,width=\columnwidth]{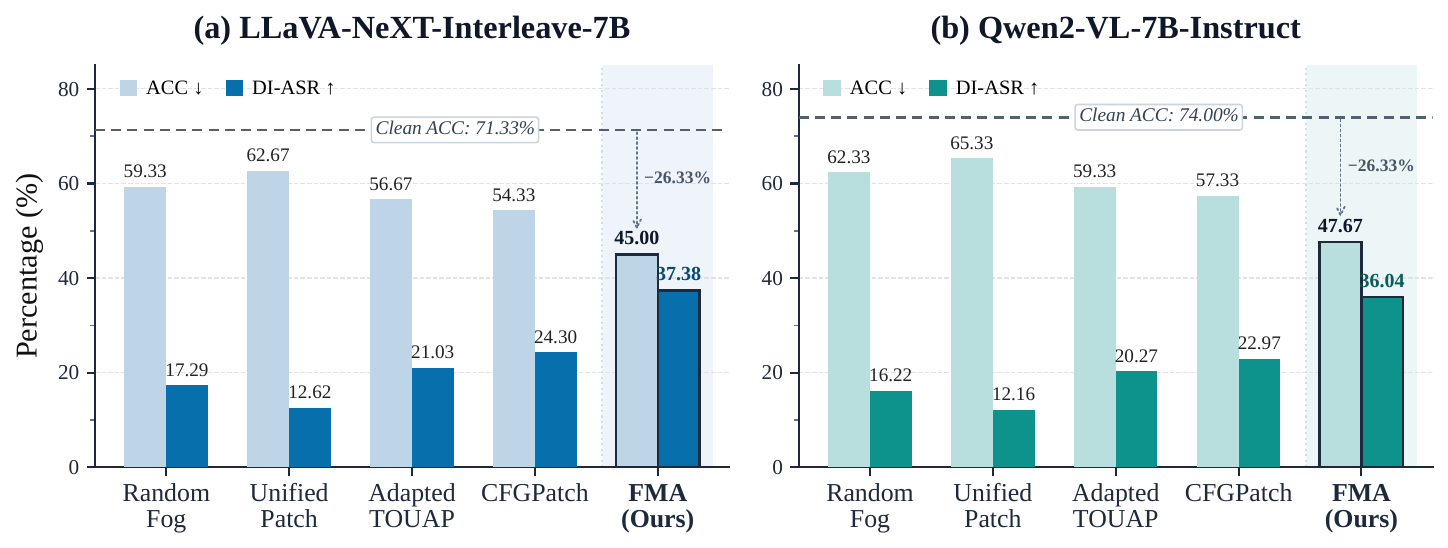}
\caption{\textbf{Dual-input VQA on controlled pairs.}
Light bars show accuracy on all 300 questions. Dark bars show DI-ASR, the clean-to-wrong rate on each model's originally correct questions (214 left, 222 right).
Dashed lines mark clean accuracy. Supplementary Table~\FmaCrossRef{tab:s14} gives full results and the source encoder.}
\label{fig:dual_input_vqa}
\end{figure}

\paragraph{A clean companion observation restores many answers.}
Figure~\ref{fig:dual_input_vqa} shows that FMA causes the largest accuracy drop among the compared attacks even when both views are available.
Accuracy falls by 26.33 percentage points on each model, with clean-to-wrong rates (DI-ASR) of 37.38\% on LLaVA-NeXT-Interleave and 36.04\% on Qwen2-VL.
The DI-ASR exceeds that of CFGPatch, the strongest comparison, by about 13 points on both models. Supplementary Table~\FmaCrossRef{tab:s14} gives the full baseline results and the source encoder.
To examine the contribution of the companion modality, we keep the selected attack fixed and replace one perturbed image with its clean counterpart.
All four conditions use the same question and reference answer.
Perturbing only one branch produces far fewer failures than perturbing both (Supplementary Table~\FmaCrossText{\FmaCrossRef{tab:s8}(c)}).

The restoration test starts with questions answered correctly with both clean views.
Both recovery rates are computed on the subset of these questions that the model answers incorrectly under paired fog.
Returning clean IR corrects nearly two-thirds of these answers; returning clean VIS corrects almost three-quarters (Supplementary Table~\FmaCrossText{\FmaCrossRef{tab:s8}(c)}).
Thus, many joint-input failures still have a usable semantic route through either clean observation.
FMA removes that route by affecting the two views together.
This directly connects the shared environmental attack to the loss of cross-modal compensation.
Figure~\ref{fig:qualitative_outputs} shows representative captions and answers before and after fog.

On real sensor pairs, paired-fog DI-ASR remains near 38\% across both models on M3FD/DSERT-RoLL and reaches 38.33\% on FLIR/FMB when transferred directly to the unseen InternVL2-8B architecture (Supplementary Table~\FmaCrossText{\FmaCrossRef{tab:s8}(d)}--\FmaCrossText{(f)}).
The table gives the four input conditions, restoration counts, and interaction estimates for these cohorts.

\subsection{Why the Shared Fog State Works}\label{sec:ablation}

We vary rendering and search across four surrogates on the controlled benchmark to identify what makes a single fog state effective in both bands (Figure~\ref{fig:controlled_ablations}).

\paragraph{Shared paths shape different spectral cues.}
Depth determines how far each viewing ray travels through the fog and haze layers.
Removing depth gives the largest Strict Joint ASR drop among the spatial variants on every encoder (9.0--11.0 percentage points), and removing layer-aware propagation lowers the mean by 4.8 points (Figure~\ref{fig:controlled_ablations}(a)).
Removing thermal path radiance lowers mean IR ASR by 11.6 points and VIS ASR by 1.2 points (Figure~\ref{fig:controlled_ablations}(b)).
On real sensor pairs, removing this term at a fixed fog state restores IR correctness in 28.4\% of original joint failures (Supplementary Sec.~\FmaCrossRef{supp:thermal}).

\paragraph{Joint scoring directs search toward paired failures.}
A single-branch objective raises ASR in the scored band and lowers it in the companion band (Figure~\ref{fig:controlled_ablations}(g)).
Scoring both branches yields 27.8\% mean Strict Joint ASR, compared with 20.4\% for VIS-only scoring and 18.6\% for IR-only scoring (Figure~\ref{fig:controlled_ablations}(c)).

\paragraph{Sharing the state improves search at matched fidelity.}
At equal budgets and similar mean LPIPS~\cite{zhang2018lpips} (0.212--0.214), the 7D search gains 4.4 joint ASR points over independent 12D fields and 3.4 over 12D fields with a shared density seed (Figure~\ref{fig:controlled_ablations}(e)).
The default 12D setting reaches 29.5\% at a higher LPIPS of 0.257 (Figure~\ref{fig:controlled_ablations}(f)).
Across the six real-sensor cells, the gain over quality-matched independent search is 3.18 points at \(B=840\) (scene-cluster 95\% CI [1.62, 4.74], Supplementary Table~\FmaCrossRef{tab:s7}).
On the real-sensor budget grid, the shared search first exceeds 25\% at 840 paired evaluations, compared with 3,360 for the quality-matched independent search (Supplementary Table~\FmaCrossRef{tab:s6}).

\paragraph{Fidelity controls limit distortion.}
Removing \(\mathcal{L}_x\) or hierarchical fidelity weighting raises joint ASR and increases mean LPIPS by 0.053 and 0.074, respectively (Figure~\ref{fig:controlled_ablations}(d)).
Full FMA lies on the observed attack--fidelity frontier (Figure~\ref{fig:controlled_ablations}(f)).
All encoders agree on the direction of each joint ASR change (Figure~\ref{fig:controlled_ablations}(h), Supplementary Table~\FmaCrossRef{tab:s11}).

\subsection{Sensor, Encoder, and Temporal Transfer}\label{sec:real}
\begin{table*}[!tbp]
\centering
\begingroup
\setlength{\fboxsep}{1pt}
\captionsetup{skip=6pt}
\caption{Strict Joint ASR (\%) on real VIS--IR pairs.
\protect\colorbox{fmaEmphasisBlue}{\strut Dark blue} and \protect\colorbox{fmaLightBlue}{\strut pale blue} mark first and second per column.
Counts and uncertainty estimates are in Supplementary Table~\FmaCrossRef{tab:s3}.}
\label{tab:m7}
\fmatablefont
\setlength{\tabcolsep}{2pt}
\renewcommand{\arraystretch}{0.92}
\newcolumntype{C}[1]{>{\centering\arraybackslash}p{#1}}
\begin{tabularx}{\linewidth}{@{}L*{3}{C{27pt}C{35pt}C{34pt}C{23pt}}}
\toprule
\multirow{2}{*}{Method} & \multicolumn{4}{c}{M3FD~\cite{liu2022tardal}} & \multicolumn{4}{c}{FLIR-Aligned~\cite{zhang2020cfr}} & \multicolumn{4}{c}{FMB~\cite{liu2023segmif}} \\
\cmidrule(lr){2-5}\cmidrule(lr){6-9}\cmidrule(lr){10-13}
& OpenAI & OpenCLIP & MetaCLIP & EVA & OpenAI & OpenCLIP & MetaCLIP & EVA & OpenAI & OpenCLIP & MetaCLIP & EVA \\
\midrule
Random Fog (matched) & 11.22 & 12.90 & 9.24 & 9.60 & 10.77 & 12.43 & 8.79 & 9.18 & 10.53 & 12.22 & 8.43 & 8.85 \\
Unified Patch~\cite{wei2023unified} & 13.78 & 17.20 & 8.15 & 12.63 & 13.33 & 16.76 & 8.24 & 12.24 & 13.16 & 16.11 & 7.87 & 11.98 \\
Adapted TOUAP~\cite{hu2025touap} & 16.84 & 20.43 & 14.67 & 18.18 & 16.41 & 20.00 & 14.29 & 17.86 & 15.79 & 19.44 & 13.48 & 17.19 \\
CFGPatch~\cite{chen2026cfgpatch} & 18.37 & 22.04 & 16.30 & \cellcolor{fmaLightBlue}19.19 & 17.95 & 21.62 & 15.38 & \cellcolor{fmaLightBlue}18.88 & 17.37 & 21.11 & 15.17 & \cellcolor{fmaLightBlue}18.23 \\
AdvHaze~\cite{gao2021advhaze} & \cellcolor{fmaLightBlue}20.41 & \cellcolor{fmaLightBlue}24.73 & \cellcolor{fmaLightBlue}16.85 & 17.68 & \cellcolor{fmaLightBlue}20.00 & \cellcolor{fmaLightBlue}23.78 & \cellcolor{fmaLightBlue}16.48 & 17.35 & \cellcolor{fmaLightBlue}19.47 & \cellcolor{fmaLightBlue}23.33 & \cellcolor{fmaLightBlue}15.73 & 16.67 \\
\specialrule{\lightrulewidth}{0pt}{0pt}
\textbf{FMA (Ours)} & \cellcolor{fmaEmphasisBlue}\textbf{26.02} & \cellcolor{fmaEmphasisBlue}\textbf{30.11} & \cellcolor{fmaEmphasisBlue}\textbf{21.20} & \cellcolor{fmaEmphasisBlue}\textbf{21.72} & \cellcolor{fmaEmphasisBlue}\textbf{25.13} & \cellcolor{fmaEmphasisBlue}\textbf{29.19} & \cellcolor{fmaEmphasisBlue}\textbf{20.33} & \cellcolor{fmaEmphasisBlue}\textbf{20.92} & \cellcolor{fmaEmphasisBlue}\textbf{24.74} & \cellcolor{fmaEmphasisBlue}\textbf{28.89} & \cellcolor{fmaEmphasisBlue}\textbf{19.66} & \cellcolor{fmaEmphasisBlue}\textbf{20.31} \\
\specialrule{\heavyrulewidth}{0pt}{0pt}
\end{tabularx}
\endgroup
\end{table*}

\paragraph{The joint failures extend to real sensor pairs.}
Across the three real datasets and four CLIP families, FMA reaches 24.02\% mean Strict Joint ASR, compared with 10.35\% for matched Random Fog (Table~\ref{tab:m7}).
The gain is 13.67 percentage points, with a scene-cluster 95\% interval of [11.84, 15.50].
The consistent improvement across datasets connects the controlled finding to registered observations acquired by different visible and thermal sensors.

Paired failure-state analysis reveals where the additional joint failures come from.
Among the new joint failures induced beyond Random Fog on the diagnostic cohort, 82.5\% begin as errors in only one branch.
FMA often turns an already asymmetric failure into a loss of both observations.
The consequence also reaches fused predictions: both logit fusion and a learned cross-attention readout fail more often under FMA than under the matched random and patch controls (Supplementary Table~\FmaCrossRef{tab:s4}).

\paragraph{Selected images remain effective beyond the source encoder.}
For each source surrogate, we freeze the selected image pairs and evaluate the other three encoder families.
Across the resulting 36 off-diagonal cells, mean Strict Joint ASR is 17.16\% for FMA and 10.12\% for Random Fog (Supplementary Table~\FmaCrossRef{tab:s9}).
The transfer result shows that the selected atmospheric changes affect semantic evidence used by multiple encoders, extending their effect beyond the scores that guided selection.

\paragraph{The atmospheric state can persist as the scene moves.}
On naturally foggy DSERT-RoLL images, FMA continues to induce more joint failures than matched Random Fog.

We also optimize one frame, freeze its atmospheric parameters, and render them with the depth of later frames.
Joint success remains 20.80\% at a five-second offset, compared with 9.60\% for the interval-matched random control (Supplementary Table~\FmaCrossRef{tab:s10}).
Thus, a selected environment can retain adversarial effect across changing views without a fresh parameter search for every frame.

\subsection{Robustness to Input Preprocessing}\label{sec:preprocessing}

We apply the same preprocessing to clean inputs, Random Fog, and FMA with the selected attacks held fixed.

\begin{table}[tbp]
\centering
\caption{\textbf{Fixed-attack preprocessing on controlled pairs.}
(a) Four-encoder means on original joint-correct sets. Retention is joint accuracy on processed clean pairs (Table~\FmaCrossRef{tab:preprocessing_classification}).
(b) Recovery is the share of 80 FMA-induced errors corrected per model. DI-ASR uses 214 and 222 clean-correct questions, respectively.}
\label{tab:preprocessing}
\begingroup\fmatablefont
\setlength{\tabcolsep}{3pt}
\renewcommand{\arraystretch}{1.04}
\textbf{(a) Classification robustness (\%)}\par\smallskip
\begin{tabularx}{\linewidth}{@{}Lrrr@{}}
\toprule
\multirow{2}{*}{Preprocessing} & \multirow{2}{*}{Clean ret.} & \multicolumn{2}{c}{Strict Joint ASR} \\
\cmidrule(l){3-4}
 & & Random Fog & FMA \\
\midrule
None & 100.00 & 10.18 & 23.60 \\
JPEG (Q75) & 98.97 & 9.67 & 22.19 \\
Gaussian (\(\sigma=1\)) & 96.53 & 10.18 & 21.28 \\
Median (\(3{\times}3\)) & 97.42 & 9.92 & 21.66 \\
\addlinespace[2pt]
VIS DCP & 97.94 & 7.22 & 16.25 \\
IR CLAHE & 97.42 & 8.12 & 18.70 \\
\rowcolor{fmaLightBlue}
VIS DCP + IR CLAHE & 95.36 & 5.80 & 13.03 \\
\bottomrule
\end{tabularx}
\par\smallskip
\textbf{(b) Dual-input VQA (\%)}\par\smallskip
\begin{tabularx}{\linewidth}{@{}Lrrrr@{}}
\toprule
\multirow{2}{*}{Preprocessing} & \multicolumn{2}{c}{LLaVA-NeXT} & \multicolumn{2}{c}{Qwen2-VL} \\
\cmidrule(lr){2-3}\cmidrule(l){4-5}
 & Recovery & DI-ASR & Recovery & DI-ASR \\
\midrule
None & 0.00 & 37.38 & 0.00 & 36.04 \\
VIS DCP & 26.25 & 28.97 & 27.50 & 27.48 \\
IR CLAHE & 20.00 & 31.31 & 21.25 & 29.73 \\
\rowcolor{fmaLightBlue}
VIS DCP + IR CLAHE & 38.75 & 25.23 & 41.25 & 23.42 \\
\bottomrule
\end{tabularx}
\endgroup
\end{table}

Even under combined modality-specific restoration (VIS DCP + IR CLAHE), FMA retains \textbf{more than double} the Strict Joint ASR of equally processed Random Fog (\textbf{13.03\% vs.\ 5.80\%}; Table~\ref{tab:preprocessing}(a)), and \textbf{over 58\%} of FMA-induced dual-input VQA errors remain unrecovered (Table~\ref{tab:preprocessing}(b)).

JPEG, Gaussian blur, and median filtering lower FMA's unprocessed Strict Joint ASR of 23.60\% by at most 2.32 percentage points.
The combined restoration uses visible dehazing (DCP~\cite{he2011dcp}) and thermal contrast enhancement (CLAHE~\cite{zuiderveld1994clahe}).
The gap between FMA and equally processed Random Fog is 7.23 points on the controlled benchmark (paired 95\% CI [4.29, 10.25], Supplementary Table~\FmaCrossRef{tab:preprocessing_intervals}).
On 1,137 initially joint-correct real-sensor pair--encoder evaluations, combined preprocessing leaves a pooled Strict Joint ASR of 13.63\% for FMA and 6.16\% for Random Fog, a gap of 7.48 points (Supplementary Table~\FmaCrossRef{tab:preprocessing_real}).

In dual-input VQA, DCP and CLAHE together recover more answers than either transform alone.
Recovery tracks the original FMA errors. DI-ASR also includes new errors on the original clean-correct set and remains near one quarter on both models (Table~\ref{tab:preprocessing}(b)).
Applying the same transforms to clean inputs lowers accuracy by 2.33 points per model (Supplementary Table~\FmaCrossRef{tab:preprocessing_vqa}).

\section{Conclusion}\label{sec:conclusion}

FMA shows how one fog environment can disrupt complementary visible and thermal observations.
Shared scene propagation and distinct spectral responses produce simultaneous failures that extend from classification to language outputs.
Restoring a clean companion image recovers many answers lost under paired fog.
These results make the common environment a concrete part of evaluating cross-spectral robustness.\hypertarget{fma.main.end}{}

\section*{LLM Usage Statement}

During the preparation of this manuscript, large language models (LLMs) were used solely as an assistive editing tool for grammar checking, typo correction, and improving the clarity and fluency of author-written text.
LLMs were not used to generate research ideas, formulate the scientific problem, derive the cross-spectral atmospheric propagation and rendering models, design the optimization algorithm, or conduct experimental executions and statistical analyses---all of which were conceived and carried out entirely by the authors.
Every LLM-assisted textual refinement was carefully reviewed and verified by the authors to ensure technical precision and fidelity to our empirical findings, and the authors take full responsibility for the entire content of this paper.
Note that the use of GPT-4o as a blinded evaluation judge for captioning and VQA scoring is an experimental protocol component rather than a writing aid, and is fully documented in Section~\ref{sec:setup} and Supplementary Section~\FmaCrossRef{supp:scoring}.

{\small
\hypersetup{urlcolor=black}
\renewcommand{\bibsection}{\section*{\Large\refname}}
\bibliographystyle{ieeenat_fullname}
\bibliography{CVPR2027_References_Optimized}
}

\end{document}